\documentclass[letterpaper]{article} % DO NOT CHANGE THIS
\usepackage[preprint]{aaai2027}  % DO NOT CHANGE THIS
\usepackage{multirow}
\usepackage[table]{xcolor}
\definecolor{ProtLingo}{HTML}{EAF6E8}
\usepackage[hyphens]{url}  % DO NOT CHANGE THIS
\usepackage{graphicx} % DO NOT CHANGE THIS
\usepackage{natbib}  % DO NOT CHANGE THIS AND DO NOT ADD ANY OPTIONS TO IT
\usepackage{caption} % DO NOT CHANGE THIS AND DO NOT ADD ANY OPTIONS TO IT
\usepackage{algorithm}
\usepackage{algorithmic}

\usepackage{amsmath}
\usepackage{amssymb}
\usepackage{amsfonts}
\usepackage{booktabs} % for better-looking tables

\title{ProtLingo: Efficient Protein Language Modeling \\
via Conditional Memory and Expert Routing}

\author{
    Mingrui Li\textsuperscript{\rm 1}\thanks{Equal contribution.},
    Sixian Shen\textsuperscript{\rm 1}\footnotemark[1],
    Minzhang Li\textsuperscript{\rm 1},
    Ruiyi Zhang\textsuperscript{\rm 1},
    Kexin Zhang\textsuperscript{\rm 1},
    Jiakai Zhang\textsuperscript{\rm 1},
    Jingyi Yu\textsuperscript{\rm 1}
}
\affiliations{
    \textsuperscript{\rm 1}ShanghaiTech University
}

\begin{document}

\maketitle

\begin{abstract} % 大约17-18行

Proteins perform diverse cellular functions, and even single amino-acid substitutions can alter stability, activity, or molecular interactions. Protein language models (PLMs) provide a scalable approach for modeling such sequence--function relationships from unlabeled sequences, but increasing the size of dense Transformer backbones often brings substantial computational cost without consistently improving mutation-sensitive prediction. We introduce ProtLingo, an efficient PLM framework that augments a pretrained single-sequence backbone with conditional local memory and sparse expert routing. ProtLingo maps contextual residue representations into route-specific discrete codes, composes centered local windows into latent $N$-gram addresses, and retrieves reusable residual signals associated with recurring local sequence contexts. In parallel, selected feed-forward blocks are upcycled into sparse Mixture-of-Experts layers with shared and routed experts, enabling residue-dependent computation while activating only a subset of parameters. Experiments on protein fitness prediction, FLIP benchmarks, and supervised contact prediction show that ProtLingo achieves competitive performance with a 150M-scale backbone, including strong parameter efficiency on mutation-effect prediction and preserved long-range structural representations. 

% Protein language models have become widely used foundations for protein structure prediction, fitness estimation, and mutation-effect analysis, yet simply scaling dense Transformer backbones faces diminishing returns in cost and downstream performance. We introduce ProtLingo, a protein language model framework that augments a pretrained PLM backbone with two conditional mechanisms for protein sequence modeling. ProtLingo uses centered latent $N$-gram memory to retrieve motif-conditioned residual features from local route-code windows, and applies MoE upcycling to convert selected dense feed-forward blocks into routed experts. This design preserves pretrained evolutionary knowledge while introducing local motif-aware specialization and conditional computation without relying on external MSAs or structural inputs. We evaluate ProtLingo on protein fitness prediction, clinical variant classification, FLIP benchmarks, supervised contact map prediction, and intrinsic language modeling. Experiments show that ProtLingo achieves competitive performance within the 150M-scale backbone regime, preserves structure-relevant contact representations, and benefits from complementary contributions of motif-conditioned memory and sparse expert routing. These results suggest that conditional memory and computation provide an effective direction for improving mutation-sensitive protein language modeling beyond dense parameter scaling.
\end{abstract}

% Motivation, gap, contributions. End with an explicit bulleted contributions list.

% Comments by zkx
% 1. 第一段要科学问题/算法问题各占一半，以科学问题引入最好
% 2. 第二段和第三段的开头对MSA过于强调，要指明现有方法的局限。
%   a. scaling的边界收益低，对于特定的任务如mutation等scaling的效果更不明显
%   b. retrival等等方式是通过evolution shortcut，不能解决evolution 无关的问题
%   c. 没有在模型架构设计上显式引入inductive bias
% 3. 修改完intro后，整体检查一下描述是否与实验结果consistent
% 4. 叙事上要高屋建瓴，要有金句，但是不要有冗余，不要有长难句

\section{Introduction}

 \begin{figure}[tp]
    \centering
    \includegraphics[width=\linewidth]{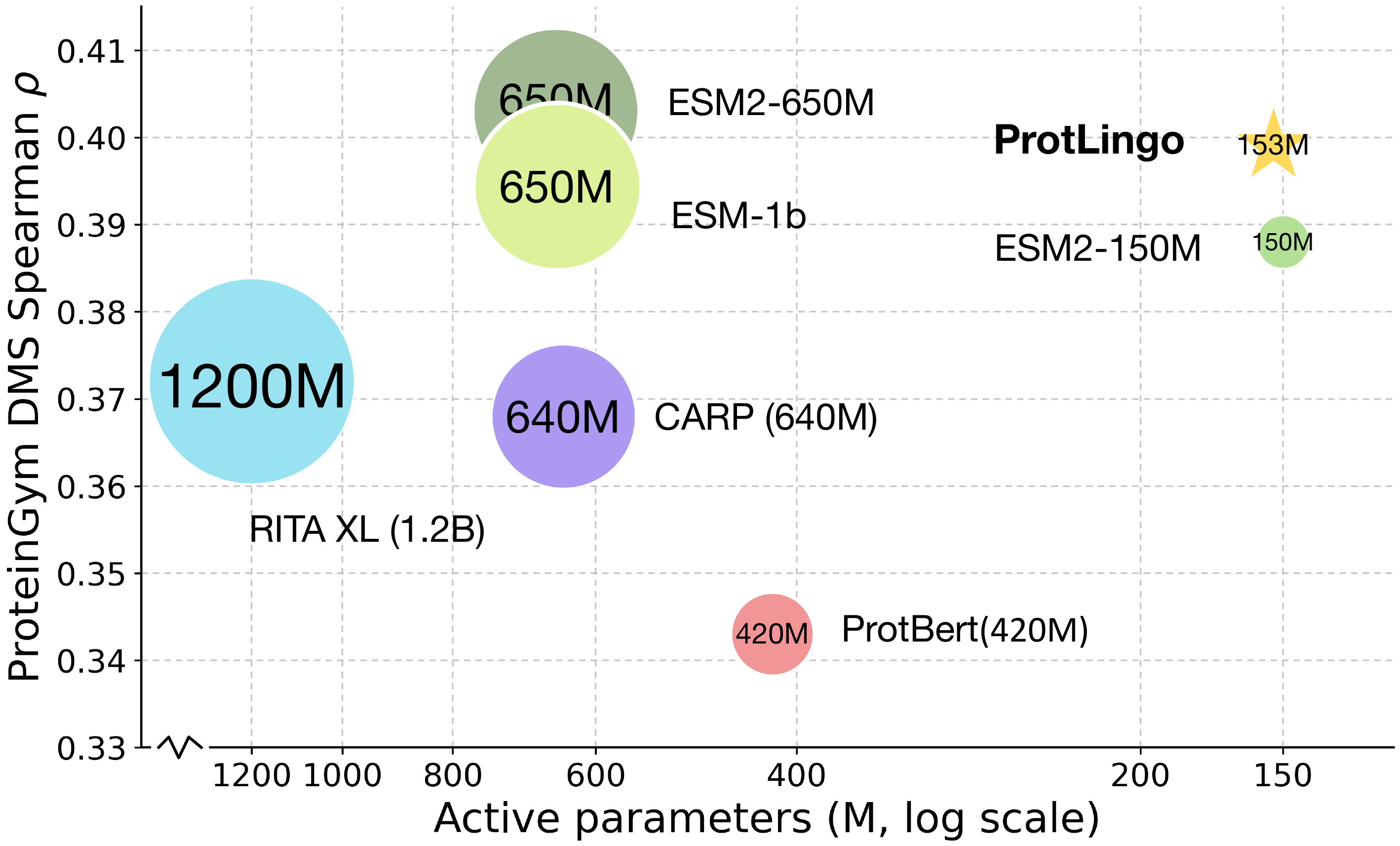}
    \caption{\textbf{Performance-parameter comparison of PLMs.} Bubble size indicates total model parameters, while the x-axis reports active parameters in millions on a log scale and the y-axis reports Spearman. ProtLingo achieves competitive DMS prediction performance with only 153M active parameters, matching or exceeding larger dense baselines such as ESM2-150M, ProtBert, CARP, and RITA XL while using fewer active parameters than ESM2-650M and ESM-1b.}
    \label{fig:motif}
  \end{figure}
Proteins perform most cellular functions, and even a single amino-acid substitution can alter their stability, activity, or interactions. Predicting these sequence–function relationships is therefore crucial for protein engineering, variant interpretation, and therapeutic discovery, yet experiments cover only a tiny fraction of the vast sequence space. Protein language models (PLMs) offer a scalable alternative by learning transferable representations from large collections of unlabeled sequences, supporting tasks from structure prediction to fitness estimation~\cite{rives2021biological,jumper2021highly,notin2023proteingym}. However, scaling dense Transformer backbones brings rapidly increasing computational costs without consistently improving downstream performance~\cite{akiyama2026expanding}. This limitation motivates architectures that use model capacity more effectively and incorporate inductive biases tailored to protein sequences~\cite{chengTrainingComputeOptimalProtein2024}.

Following this direction, prior work has enriched PLMs with evolutionary information from multiple sequence alignments~\cite{rao2021msa,eddy2011accelerated}, geometric information from protein structures~\cite{su2023saprot}, or related sequences retrieved from external databases~\cite{notin2022tranception}. Although these approaches improve biological grounding, they depend on homologous sequences, structural inputs, or external resources whose availability and quality vary across proteins. They also leave open a complementary question: how can a single-sequence PLM better exploit the local context already present within its input? This question is especially important for mutation-sensitive prediction, because the effect of a substitution often depends on nearby conserved residues, functional sites, and local residue environments~\cite{ng2003sift,bogan1998anatomy,porter2004catalytic}. Standard dense PLMs can model such dependencies implicitly, but provide no explicit mechanism for reusing recurring local contexts or adapting computation across residues. An effective architecture should therefore preserve broad pretrained knowledge while introducing local-context memory and residue-dependent computation.

We present ProtLingo, an efficient protein language modeling framework that augments a pretrained dense backbone with conditional memory and expert routing. Its conditional memory maps contextual residue representations into route-specific discrete codes and composes codes from centered local windows into latent \(N\)-gram addresses. These addresses retrieve reusable residual signals associated with recurring local sequence contexts. The retrieved memories are compressed and locally refined before being injected into selected backbone layers, allowing ProtLingo to adapt residue representations without overwriting the general knowledge encoded by pretraining. ProtLingo further upcycles selected dense feed-forward blocks into sparse Mixture-of-Experts layers containing a shared expert and multiple routed experts. A top-1 router assigns each residue to a context-dependent expert while activating only a subset of the available parameters. Both the memory pathway and the routed experts are initialized to preserve the pretrained model function, enabling stable continuation training and gradual specialization. Together, conditional memory and expert routing allow ProtLingo to reuse recurring local contexts and allocate computation adaptively across residue environments, improving mutation-sensitive sequence modeling without relying on substantially larger dense backbones.

We evaluate ProtLingo on protein fitness prediction, clinical variant classification, FLIP benchmarks and supervised contact prediction. Across these tasks, ProtLingo achieves competitive performance with a 150M-scale backbone, demonstrating that conditional memory and expert routing can improve sequence modeling without substantially enlarging the dense model. On supervised contact prediction, ProtLingo retains precision comparable to the ESM2-150M backbone, suggesting that continuation training preserves structure-relevant long-range residue representations. Ablation studies further examine centered latent $N$-gram memory and MoE upcycling. The results show that local-context memory and routed experts provide complementary gains. We will release the source code and model checkpoints upon publication.

\section{Related Work}
  \begin{figure*}[t]
    \centering
    \includegraphics[width=0.9\textwidth]{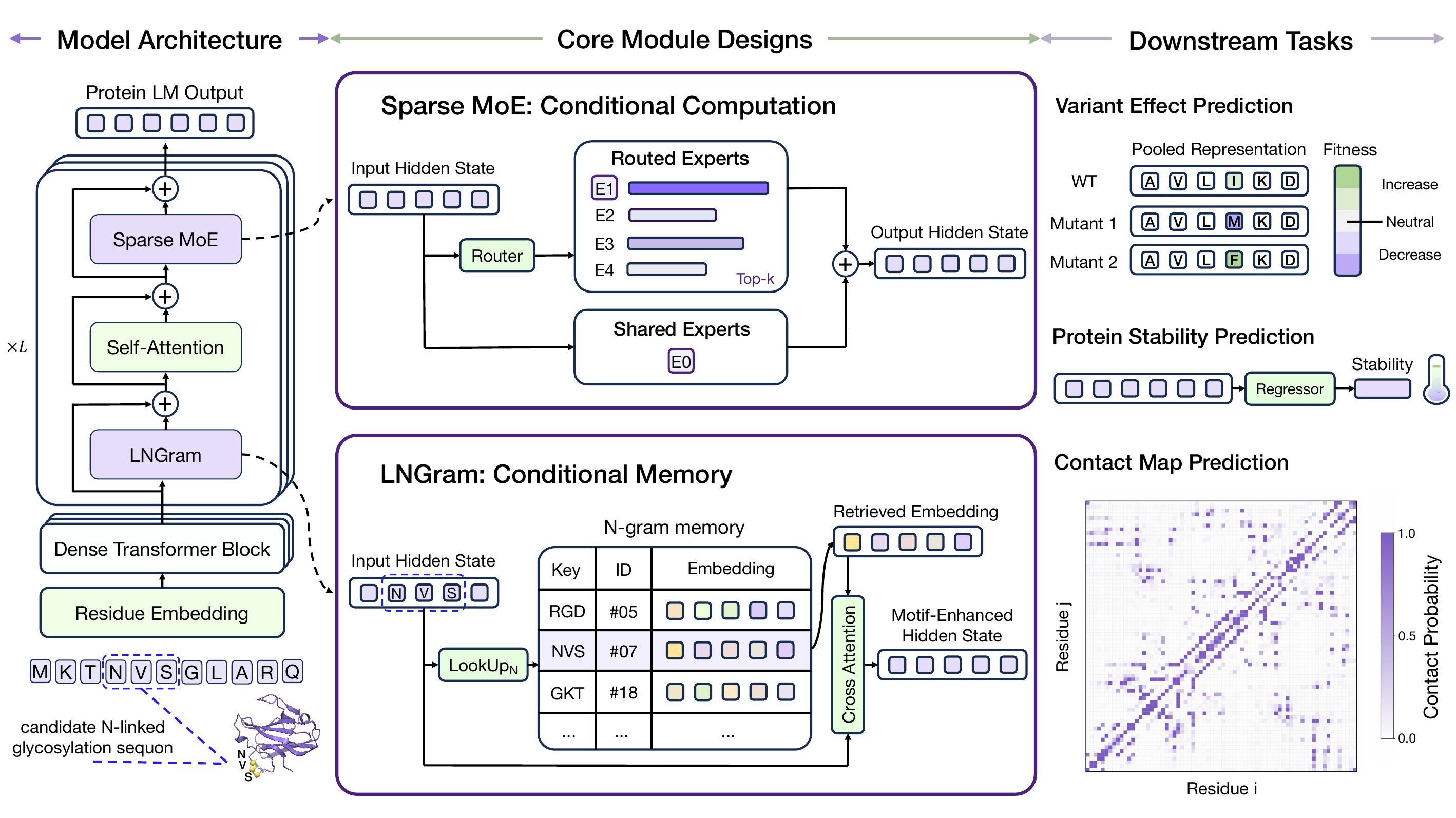}
    \caption{\textbf{Overview of the ProtLingo architecture and downstream applications.} ProtLingo augments a dense protein Transformer backbone with two conditional modules:
a Sparse MoE layer for sparse expert routing and an LNGram module for motif-aware conditional memory retrieval.
Given protein sequences and candidate motifs such as N-linked glycosylation sequons, ProtLingo produces enriched protein representations for variant effect prediction, protein stability prediction, and contact map prediction.
}
    \label{fig:method}
  \end{figure*}
\subsection{Protein Language Models}

PLMs learn transferable representations from large unlabeled sequence corpora via self-supervised pretraining. Early distributed representations~\citep{asgari2015continuous,asgari2019dimotif,heinzingerModelingAspectsLanguage2019,raoEvaluatingProteinTransfer2019} established that sequence modeling captures meaningful biological signal, and Transformer-based pretraining consolidated this paradigm at scale: masked (bidirectional) models such as ProtTrans~\citep{elnaggarProtTransCrackingLanguage2021}, ESM-1b~\citep{rives2021biological}, ESM2~\citep{linEvolutionaryscalePredictionAtomiclevel2023}, and XTrimoPGLM~\citep{chenXTrimoPGLMUnified100BScale2024} learn contextual residue representations that encode conservation, structure, and contacts, enabling zero-shot variant-effect prediction~\citep{yangADeLAAutomaticDense2021}, while autoregressive models~\citep{hesslowRITAStudyScaling2022,nijkampProGen2ExploringBoundaries2022} extend PLMs to controllable sequence generation with experimentally validated designs~\citep{madaniLargeLanguageModels2023}. Subsequent work has broadened the paradigm to structure- and function-conditioned modeling~\citep{hayesSimulating500Million2025,wangDiffusionLanguageModels2024} and improved its efficiency through data curation and training recipes~\citep{fournierProteinLanguageModels2024,notin2023proteingym,chengTrainingComputeOptimalProtein2024}. At the same time, systematic evaluations indicate that downstream gains are not monotone in model size, with variant-effect performance saturating beyond mid-scale regimes~\citep{chengTrainingComputeOptimalProtein2024,notin2023proteingym}. 
% Common to these models is a uniform computation principle: every residue is processed through the same parameter pathway, and capacity grows only by scaling that pathway as a whole.

\subsection{Conditional Memory for Local Sequence Patterns}

PLMs encode sequence regularities implicitly within a shared parametric backbone. While such distributed representations are flexible, they offer limited explicit mechanisms for reusing a recurring local pattern when the same or related context appears in another sequence. This issue is particularly relevant to proteins, where local residue contexts, conserved motifs, and short functional segments often recur across evolutionarily or functionally related sequences. Earlier protein representations captured such regularities through fixed-length $k$-mers or learned variable-length segments~\citep{asgari2015continuous,asgari2019dimotif}, but these units are typically defined before, or outside, the contextual encoder. Conditional memory provides a complementary way to increase model capacity by storing reusable patterns in addressable tables and retrieving entries conditioned on the current input. Engram constructs memory addresses from explicit token $n$-grams, whereas LNGram derives discrete addresses from latent states, enabling context-dependent retrieval without relying on a predefined phrase vocabulary~\citep{cheng2026engram,zheng2026lngram}. 
% However, existing conditional-memory methods are not specifically designed for the local, bidirectional representations used by masked protein encoders. This leaves open how addressable memory can be adapted to represent reusable local sequence patterns in PLMs.

 \subsection{Mixture of Experts}

Dense PLMs process every residue through the same feed-forward parameter set, even when residue contexts differ substantially. Increasing this uniformly activated capacity improves model expressiveness, but it also increases computation for every token. Conditional computation offers an alternative by decoupling total model capacity from per-token active computation, selecting input-dependent parameter pathways instead of activating the full model for every residue. MoE models instantiate this idea through sparse routing, where each token is assigned to a small subset of experts~\citep{fedus2022switch,zoph2022stmoe}. Sparse upcycling further adapts this principle to pretrained dense models by reusing learned feed-forward parameters rather than training a sparse architecture from scratch~\citep{komatsuzaki2023sparse}. In protein modeling, sparse expert architectures have shown that model capacity can be expanded while retaining competitive predictive performance~\citep{sun2024aido}. 
% However, existing studies primarily treat routing as a scaling mechanism, leaving less explored whether expert pathways organize around biologically meaningful patterns such as motifs, residue contexts, or mutation-sensitive functional constraints.

\section{Method}

\subsection{Overview}
\label{sec:method-overview}

As shown in Figure~\ref{fig:method}, ProtLingo adapts a pretrained protein language model through two complementary conditioning mechanisms: centered latent $N$-gram memory and sparse expert routing.
The memory module retrieves motif-conditioned residuals using route-specific discrete addresses derived from centered residue neighborhoods.
MoE upcycling converts selected dense feed-forward blocks into one shared and multiple routed experts, providing context-dependent computation while reusing pretrained weights.
Together, these modules retain the evolutionary and structural knowledge of the pretrained backbone while adding compute-efficient specialization for mutation-sensitive sequence modeling.

ProtLingo retains the backbone representation pipeline (Section~\ref{sec:esm2-backbone}), inserts centered latent $N$-gram memory at selected layers (Section~\ref{sec:lngram-memory}), upcycles selected feed-forward blocks into sparse MoE layers (Section~\ref{sec:moe-upcycling}), and uses continuation pretraining with auxiliary routing losses (Section~\ref{sec:training-objective}).

\subsection{Model Backbone and Layer-wise Adaptation}
\label{sec:esm2-backbone}

ProtLingo builds on a pretrained masked protein language model.
Given a protein sequence $x=(x_1,\ldots,x_L)$, the backbone produces hidden states
\begin{equation}
    H^{(0)}, H^{(1)}, \ldots, H^{(B)},
\end{equation}
where $H^{(0)}\in\mathbb{R}^{L\times d}$ is the embedding output,
$H^{(\ell)}\in\mathbb{R}^{L\times d}$ is the representation after $\ell$ Transformer blocks, and $B$ is the number of blocks.
Centered latent $N$-gram memory is inserted before selected blocks, while MoE upcycling replaces selected dense feed-forward blocks with sparse experts.
This design introduces conditional memory and computation while retaining the pretrained backbone.

\subsection{Centered Latent $N$-gram Memory}
\label{sec:lngram-memory}

Centered latent $N$-gram memory provides conditional memory for local motif adaptation, where a residue's function depends on both its identity and surrounding context.
At a selected insertion layer, let $H\in\mathbb{R}^{L\times d}$ denote the contextual residue representations; the layer index is omitted for clarity.
For each residue $i\in\{1,\ldots,L\}$, ProtLingo constructs the ordered centered window
\begin{equation}
    \mathcal{W}_i
    =
    (i-s,\ldots,i,\ldots,i+s),
\end{equation}
where $N=2s+1$ is the window size.
This window captures local context relevant to motif- and mutation-sensitive modeling.

Rather than indexing raw amino-acid $N$-grams, ProtLingo constructs memory addresses from contextual latent route codes.
Let $h_i=H_{i,:}\in\mathbb{R}^{d}$ denote the representation of residue $i$.
Each representation is normalized and projected into route logits:
\begin{equation}
    \bar h_i
    =
    \mathrm{RMSNorm}(h_i),
    \qquad
    q_i
    =
    W_q\bar h_i
    \in\mathbb{R}^{d}.
\end{equation}
The logits are reshaped into $R=d/b$ routes,
$q_{i,r}\in\mathbb{R}^{b}$ for $r\in\{1,\ldots,R\}$, where $b$ is the configurable number of bits per route.
Each route is thresholded and binary-encoded as
\begin{equation}
    c_{i,r}
    =
    1+\sum_{j=1}^{b}
    2^{j-1}\mathbb{I}[q_{i,r,j}>0]
    \in\{1,\ldots,2^b\}.
\end{equation}
Thus, each residue produces $R$ route-specific symbols rather than one global code.
Out-of-boundary positions use the reserved symbol $0$, preventing collisions with valid residue symbols.

For each residue $i$ and route $r$, ProtLingo combines the route symbols within $\mathcal{W}_i$ into a centered latent $N$-gram address:
\begin{equation}
    a_{i,r}
    =
    \mathrm{addr}
    \left(
        c_{i-s,r},\ldots,c_{i,r},\ldots,c_{i+s,r}
    \right).
\end{equation}
Each address retrieves a vector from its route-specific memory table:
\begin{equation}
    m_{i,r}
    =
    \mathcal{T}_r[a_{i,r}]
    \in\mathbb{R}^{d_m},
\end{equation}
where $\mathcal{T}_r$ is the learnable table for route $r$ and $d_m$ is the per-route memory dimension.
The route-wise memory vectors are concatenated:
\begin{equation}
    m_i
    =
    \mathrm{Concat}
    \left(
        m_{i,1},\ldots,m_{i,R}
    \right)
    \in\mathbb{R}^{R d_m}.
\end{equation}
This multi-route design allows different route-logit subspaces to retrieve complementary motif-conditioned signals from the same residue neighborhood.

Because $m_i$ can be high-dimensional, ProtLingo compresses it into a latent representation:
\begin{equation}
    z_i
    =
    W_m m_i
    \in\mathbb{R}^{d_z},
\end{equation}
where $d_z$ is the latent dimension; For example, $R=320$ and $d_m=8$, the 2560-dimensional memory is compressed to $d_z=256$.
It is then projected into $k_i,v_i\in\mathbb{R}^{d}$:
\begin{equation}
    k_i=\mathrm{RMSNorm}(W_k z_i),
    v_i=\mathrm{RMSNorm}(W_v z_i).
\end{equation}
A scalar gate controls the memory contribution using the similarity between the normalized residue representation and retrieved key:
\begin{equation}
    g_i
    =
    \sigma\!\left(
        \frac{\langle \bar h_i,k_i\rangle}{\sqrt{d}}
    \right)
    \in(0,1).
\end{equation}
The gated value is refined by a lightweight local operator:
\begin{equation}
    u_i=g_i v_i,
    \qquad
    \delta_i
    =
    u_i+\mathrm{SiLU}\!\left(\mathrm{DWConv}(u)_i\right).
\end{equation}
Here, $\mathrm{DWConv}$ is a learnable causal depthwise convolution over the gated value sequence, with receptive field $(i-2,i-1,i)$.
The refined memory residual is injected before the selected Transformer block:
\begin{equation}
    \widetilde{H}
    =
    H+\alpha\Delta,
\end{equation}
where $\Delta_{i,:}=\delta_i$ and each insertion layer has its own learnable scale $\alpha$.
In the main configuration, each $\alpha$ is initialized to $1.0$.
Despite this nonzero scale, the Lngram insertion initially preserves the pretrained backbone output because the memory table and depthwise convolution weights are zero-initialized, yielding $\Delta=0$.
ProtLingo therefore starts from the pretrained function while learning motif-conditioned residual corrections through the memory pathway.

\paragraph{Surrogate-gradient Path.}
The retrieved memory entries and latent compression layer are trained through standard backpropagation, but discrete route addresses block gradients to the route projection $W_q$.
ProtLingo therefore applies a local counterfactual surrogate to each route bit: the model flips that bit while holding all others fixed and measures the resulting change in the retrieved memory representation. This difference provides a sensitivity signal for the route logits and $W_q$.
The surrogate thus trains the pre-lookup discrete addressing mechanism rather than acting as a straight-through estimator for the memory tables.

\subsection{Sparse Expert Adaptation via MoE Upcycling}
\label{sec:moe-upcycling}

To increase conditional modeling capacity, ProtLingo replaces selected dense feed-forward blocks with sparse MoE layers.
Each selected block is upcycled into one shared expert and multiple routed experts, initialized from complementary parts of the pretrained dense MLP weights, with optional small perturbations for non-primary routed experts.
This preserves the pretrained computation at initialization while allowing expert specialization during continuation training.

For residue $i$, let $U_i$ denote the input to the feed-forward block.
The router computes logits over the $E$ routed experts:
\begin{equation}
    \boldsymbol{\eta}_i
    =
    W_r\mathrm{LayerNorm}(U_i)
    \in \mathbb{R}^{E}.
\end{equation}
ProtLingo uses top-1 routing in the main configuration:
\begin{equation}
    e_i
    =
    \arg\max_{e\in\{1,\ldots,E\}}
    \left[\mathrm{softmax}(\boldsymbol{\eta}_i)\right]_e .
\end{equation}
The MoE output combines the shared expert with the selected routed expert:
\begin{equation}
    Y_i
    =
    \mathcal{E}_{\mathrm{sh}}(\bar U_i)
    +
    \mathcal{E}_{e_i}(\bar U_i),
\end{equation}
where $\bar U_i=\mathrm{LayerNorm}(U_i)$, and
$\mathcal{E}_{\mathrm{sh}}$ and $\mathcal{E}_{e_i}$ denote the shared and selected routed experts, respectively.
The selected expert is unscaled in the forward pass, while its probability provides a surrogate gradient for context-dependent selection.
To stabilize expert usage, the training objective includes a load-balancing auxiliary loss and a router z-loss:
\begin{equation}
    \mathcal{L}_{\mathrm{MoE}}
    =
    \lambda_{\mathrm{aux}}\mathcal{L}_{\mathrm{aux}}
    +
    \lambda_z\mathcal{L}_z .
\end{equation}
The auxiliary loss balances token assignment across experts, while the z-loss regularizes router logits.

\begin{figure*}[t]
  \centering
  \includegraphics[width=0.85\textwidth]{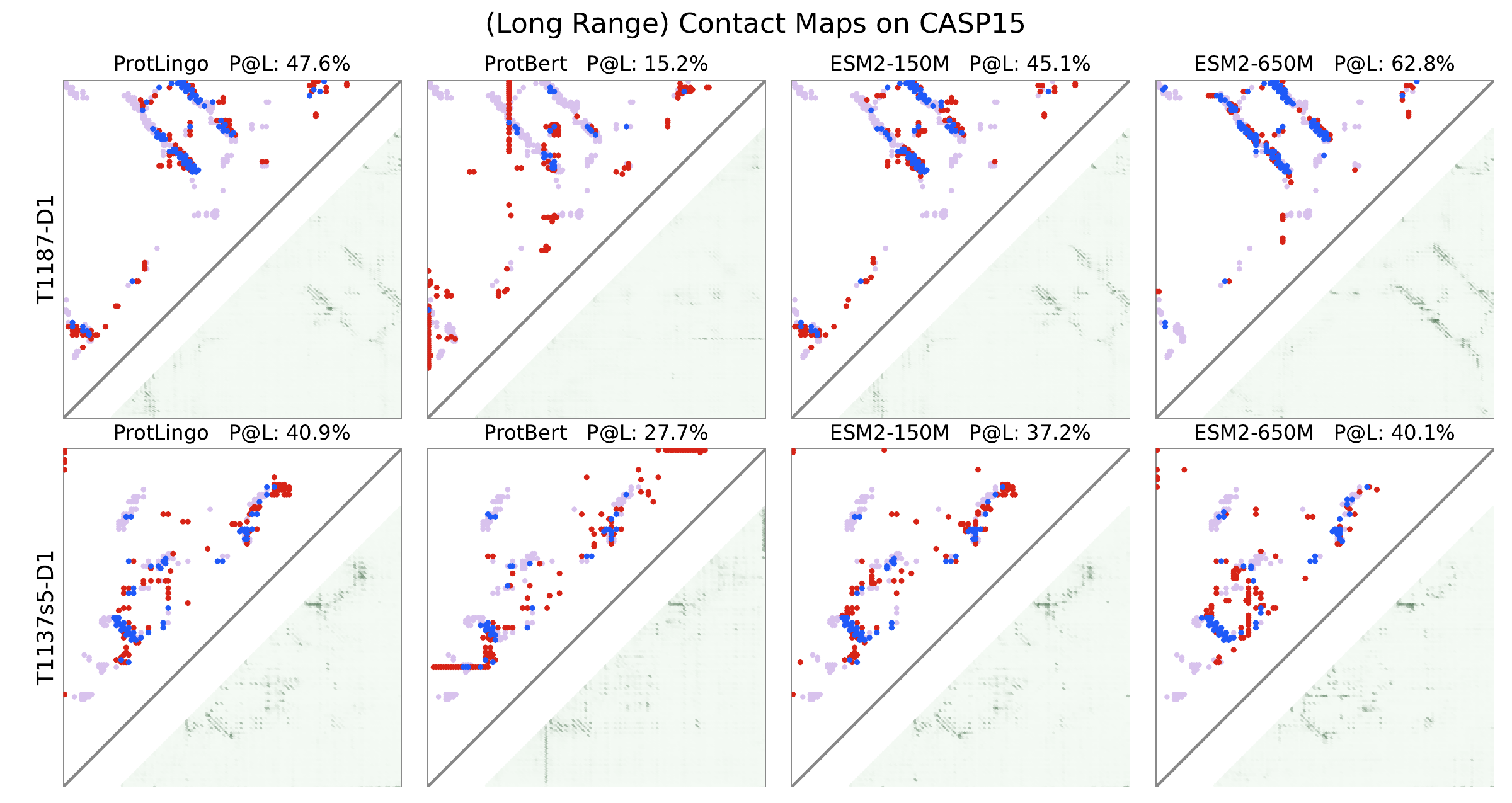}
  \caption{
  \textbf{Contact map prediction comparison on CASP15 targets.}
  We visualize predicted long-range contact maps from different pretrained encoders under the same supervised contact-head protocol.
  The comparison illustrates whether each model preserves structure-relevant nonlocal residue couplings after pretraining or adaptation.
  }
  \label{fig:contact_map_comparison}
\end{figure*}

\subsection{Training Objective}
\label{sec:training-objective}

ProtLingo continues pretraining from an ESM2 checkpoint using the masked language modeling objective:
\begin{equation}
    \mathcal{L}_{\mathrm{MLM}}
    =
    -\frac{1}{|\mathcal{M}|}
    \sum_{i\in\mathcal{M}}
    \log p_{\theta}(x_i\mid\tilde{x}),
\end{equation}
where $\mathcal{M}$ is the set of masked residue positions and $\tilde{x}$ is the corrupted input sequence.
When MoE adaptation is enabled, the final objective is
\begin{equation}
    \mathcal{L}
    =
    \mathcal{L}_{\mathrm{MLM}}
    +
    \mathcal{L}_{\mathrm{MoE}}.
\end{equation}
The MLM objective jointly updates the backbone and memory parameters. The memory table can use a separate learning-rate multiplier, allowing it to adapt faster than the pretrained backbone.

\section{Experiments}

\paragraph{Implementation Details.}
\label{sec:implementation}
ProtLingo augments selected Transformer layers with two conditional adaptation modules.
In our implementation, the centered latent $N$-gram memory is inserted into layers 1, 11, and 21 with window size $N=3$, 2 route bits, memory dimension 8, and a residual scale initialized to 1.0.
MoE upcycling is applied to layers 3--29 by converting each selected dense feed-forward block into one shared expert and four route experts initialized from the pretrained weights, with top-$K$ routing where $K=1$.
More implementation details are provided in the supplementary material.

\paragraph{Training Dataset.}
ProtLingo is trained on the same large-scale sequence source as ESM.
The training corpus is sampled from UniRef, with 80\% from UniRef50 containing approximately 65 million clustered sequences and 20\% from UniRef90 containing approximately 187 million sequences.
All sequences are processed with the ESM tokenization and filtering pipeline.

\begin{table*}[t]
  \centering
  \small
  \setlength{\tabcolsep}{3.0pt}
  \renewcommand{\arraystretch}{1.08}
  \caption{
  Fitness and mutation-effect prediction results across benchmarks.
  Best results are shown in bold and second-best results are underlined among models evaluated under our unified setting.
  Rows marked with $\dagger$ are reference baselines reported by ProteinGym for DMS prediction.
  FLIP entries marked with $\ddagger$ are literature-reported reference numbers collected from prior FLIP-style evaluations and are included for context rather than as a controlled head-to-head comparison.
  Efficiency is measured as ProteinGym DMS Spearman per billion active parameters, highlighting ProtLingo's parameter-efficient performance rather than claiming universal state-of-the-art accuracy on every benchmark.
  }
  \label{tab:fitness}
  \begin{tabular}{lccccccc}
    \hline
    \multirow{2}{*}{Model}
    & \multirow{2}{*}{Active Params}
    & \multirow{2}{*}{Efficiency}
    & \multicolumn{2}{c}{ProteinGym}
    & \multicolumn{3}{c}{FLIP} \\
    \cline{4-8}
    & & & Spearman & AUC & GB1 & AAV & Meltome \\
    \hline
    $\dagger$ RITA XL (1.2B)
    & 1.2B & 0.31 & 0.373 & 0.708 & -- & -- & -- \\
    $\dagger$ CARP (640M)
    & 640M & 0.58 & 0.369 & 0.701
    & 0.555$^{\ddagger}$ & 0.670$^{\ddagger}$ & 0.530$^{\ddagger}$ \\
    $\dagger$ ESM-1b (650M)
    & 650M & 0.61 & 0.394 & 0.719
    & 0.340$^{\ddagger}$ & 0.360$^{\ddagger}$ & 0.710$^{\ddagger}$ \\
    ESM2-150M
    & 150M & \underline{2.59} & 0.388 & 0.715 & 0.523 & 0.555 & 0.651 \\
    ESM2-650M
    & 650M & 0.62 & \textbf{0.403} & \textbf{0.723}
    & \textbf{0.555} & \underline{0.606} & \textbf{0.676} \\
    ProtBert
    & 420M & 0.82 & 0.343 & 0.688 & 0.539 & \textbf{0.608} & \underline{0.658} \\
    \cellcolor{ProtLingo}\textbf{ProtLingo}
    & \cellcolor{ProtLingo}153M
    & \cellcolor{ProtLingo}\textbf{2.61}
    & \cellcolor{ProtLingo}\underline{0.399}
    & \cellcolor{ProtLingo}\underline{0.720}
    & \cellcolor{ProtLingo}\underline{0.544}
    & \cellcolor{ProtLingo}0.576
    & \cellcolor{ProtLingo}0.650 \\
    \hline
  \end{tabular}
\end{table*}

\paragraph{Metrics.}
We use task-specific metrics for evaluation.
Contact map prediction is measured by long-range Top-$L$ precision, including P@$L$, P@$L/2$, and P@$L/5$.
Fitness prediction and mutation-effect estimation are evaluated using ProteinGym~\cite{notin2023proteingym} DMS Spearman correlation and AUC, together with ~\cite{dallago2021flip} scores on GB1, AAV, and Meltome.
Intrinsic language modeling is measured by MLM loss and perplexity.
All internally evaluated models use identical scoring scripts and evaluation splits.

\subsection{Supervised Contact Map Prediction}
\label{exp:contact}

Long-range contact prediction evaluates whether protein representations preserve nonlocal residue couplings that are important for three-dimensional folding.
Following \citet{rao2021msa}, we freeze each pretrained encoder and train the same lightweight contact head on pairwise residue features.
Contacts are defined by $\mathrm{C}\beta$--$\mathrm{C}\beta$ distance below 8\,\AA, with $\mathrm{C}\alpha$ used for glycine, and we report long-range precision at Top-$L$, Top-$L/2$, and Top-$L/5$ using residue pairs with $|i-j|\geq24$.

Table~\ref{tab:contact} reports supervised long-range contact prediction on the selected CASP15 subset with 21 domains.
The 650M-scale ESM2 baseline obtains the strongest absolute performance under this protocol.
ProtLingo achieves P@$L$=0.485, P@$L/2$=0.657, and P@$L/5$=0.816 with only 153M active parameters, slightly improving over ESM2-150M at P@$L$ and matching it at P@$L/2$.
These results suggest that the conditional memory and computation modules preserve structure-relevant representations while keeping the active-parameter budget close to the 150M-scale backbone.
We use this experiment as a representation-preservation diagnostic rather than as a claim of state-of-the-art contact prediction accuracy.

\subsection{Protein Fitness Prediction}
\label{exp:fitness}
\begin{table}[t]
  \centering
  \small
  \setlength{\tabcolsep}{4pt}
  \renewcommand{\arraystretch}{1.05}
  \caption{
  Supervised long-range contact prediction on the CASP15 subset.
  Best results are shown in bold and second-best results are underlined.
  For the 650M-scale baseline, we report the NVIDIA implementation results under the same model family.
  ProtLingo achieves competitive contact prediction performance with a compact active-parameter budget, and the comparison is intended to assess parameter-efficient representation quality rather than to claim universal state-of-the-art contact prediction accuracy.
  }
  \label{tab:contact}
  \begin{tabular}{lcccc}
    \hline
    Model & Active Params & P@$L$ & P@$L/2$ & P@$L/5$ \\
    \hline
    ESM2-150M
      & 150M & 0.484 & \underline{0.657} & \underline{0.830} \\
    ESM2-650M
      & 650M & \textbf{0.514} & \textbf{0.682} & \textbf{0.845} \\
    ProtBert
      & 420M & 0.281 & 0.353 & 0.435 \\
    \cellcolor{ProtLingo}\textbf{ProtLingo}
      & \cellcolor{ProtLingo}153M
      & \cellcolor{ProtLingo}\underline{0.485}
      & \cellcolor{ProtLingo}\underline{0.657}
      & \cellcolor{ProtLingo}0.816 \\
    \hline
  \end{tabular}
\end{table}
We evaluate mutation-sensitive protein function prediction on ProteinGym and FLIP benchmarks.
For ProteinGym deep mutational scanning assays, we follow masked-marginal scoring and compute the log-likelihood difference between mutant and wild-type residues at the mutated position.
We report Spearman correlation for continuous DMS fitness measurements and AUC for binary mutation-effect labels.
For FLIP, we report task scores on GB1, AAV, and Meltome using the same scoring protocol for internally evaluated models.

Table~\ref{tab:fitness} summarizes the results.
Among models evaluated under our unified setting, larger dense encoders generally achieve the strongest absolute scores, while ProtLingo attains the highest parameter-efficiency score.
Here efficiency is defined as ProteinGym DMS Spearman divided by active parameters in billions.
ProtLingo improves over ESM2-150M on ProteinGym Spearman and AUC, and remains close to the 150M-scale baseline on FLIP tasks.
The $\dagger$ and $\ddagger$ rows provide external reference points from ProteinGym and prior FLIP-style evaluations; they are included for context and are not treated as controlled head-to-head comparisons.

\subsection{Ablation Study}

We compare full ProtLingo with variants that remove the Sparse MoE module, the motif-aware Lngram memory module, or both.
We evaluate whether the gains come primarily from conditional computation, motif-aware memory retrieval, or their interaction.
We report changes in ProteinGym DMS Spearman correlation ($\Delta\rho$) and language-model perplexity ($\Delta$PPL) relative to full ProtLingo, where higher $\Delta\rho$ and lower $\Delta$PPL indicate better performance.

\begin{table}[t]
  \centering
  \small
  \setlength{\tabcolsep}{5pt}
  \renewcommand{\arraystretch}{1.08}
  \caption{
  Ablation study relative to full ProtLingo.
  $\Delta$ denotes the ablated-model value minus the full-model value.
  }
  \label{tab:ablation}
  \begin{tabular}{lcc}
    \hline
    Variant & PG-DMS $\Delta\rho \uparrow$
            & $\Delta$PPL $\downarrow$ \\
    \hline
    Full ProtLingo
      & -- & -- \\
    $-$  MoE
      & -0.093 & +2.461 \\
    $-$ Lngram
      & -0.357 & +6.003 \\
    $-$ MoE $-$ Lngram
      & -0.369 & +6.035 \\
    \hline
  \end{tabular}
\end{table}

\section{Model Analysis}

\begin{figure*}[t]
    \centering
    \includegraphics[width=0.85\textwidth]{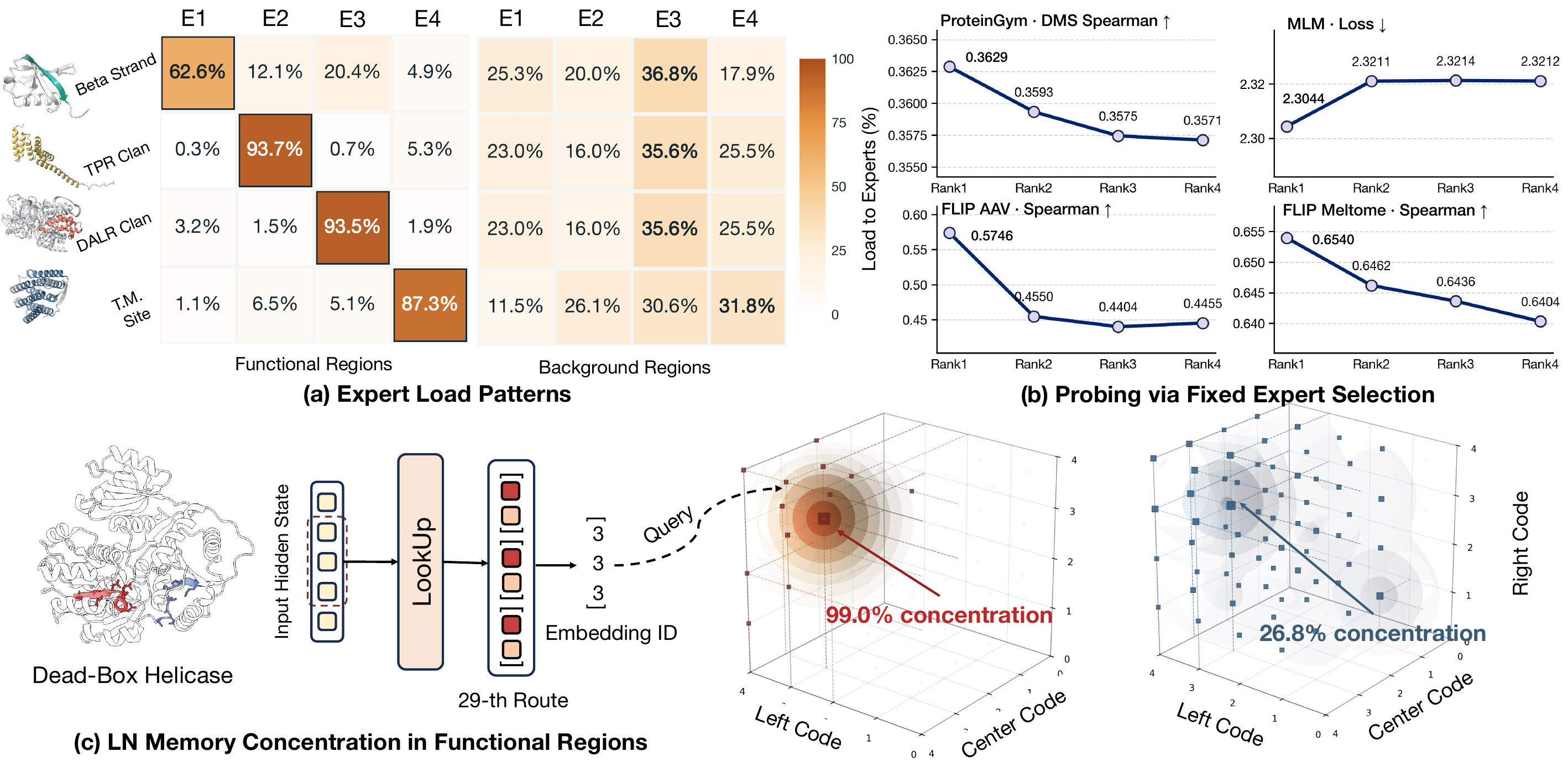}
    \caption{
    Biological specialization of sparse experts and latent $N$-gram memory in ProtLingo.
    \textbf{(a)} Expert loads for representative structural and functional annotations and same-protein, amino-acid- and position-matched backgrounds, revealing category-specific preferences.
    \textbf{(b)} Performance under Rank-1 to Rank-4 expert selection; lower-ranked alternatives reduce ProteinGym and FLIP correlations and increase MLM loss.
    \textbf{(c)} Layer-21 memory addressing for DEAD-box helicase motifs in a representative route. Centered codes show 99.0\% motif concentration versus 26.8\% for matched non-motif controls.
    }
    \label{fig:model-analysis}
\end{figure*}

% \subsection{ProtLingo Learns Motif-Aware Latent Memory}
\subsection{Motif-Aware Representations in LNgram Memory}
To test whether centered LNgram captures biologically meaningful motif representations, we asked whether (i) motifs from the same family share similar discrete route addresses and continuous residuals, (ii) motifs exhibit greater address concentration than matched non-motif one, and (iii) non-conservative or motif-localized substitutions induce greater residual changes than conservative or outside-motif controls.

We first analyzed 10,573 curated occurrences from 10 PROSITE families. The results showed that motifs from the same family formed similar route-local address combinations while retaining protein-context variation. At layer 21, the mean residual cosine similarity was 0.968 within families versus 0.756 between families. We then compared 10,447 motifs with amino-acid-composition-matched non-motif windows from the same proteins. At layer 21, motif address agreement reached 54.2\%, compared with 23.4\% for controls; Fig.~\ref{fig:model-analysis}c illustrates this concentration in a representative route. Finally, across 21,958 functional anchors and 18,290 identical-substitution controls outside motifs, non-conservative mutations caused 17.6–21.1\% greater residual perturbation than conservative mutations. Motif mutations also caused 16.8–22.8\% greater perturbation than identical substitutions outside motifs, measured as the mean local L2 distance between scaled wild-type and mutant residuals.

Together, these results show that centered Lngram learns stable, motif-aware representations that cannot be explained by protein context or amino-acid composition alone and are selectively sensitive to motif perturbation.

\subsection{Biological Specialization of Sparse Experts}
We assessed MoE routing and specialization across annotated residues. From the complete Swiss-Prot collection, we retained 461,859 length-filtered, sequence-deduplicated proteins (156,754,291 residues), yielding 4.23 billion Top-1 decisions across 27 MoE layers. Mean normalized assignment entropy was 0.971, corresponding to an effective 3.85 of four experts and indicating no global expert collapse.

  % \begin{figure*}[t]
  %   \centering
  %   \includegraphics[width=0.95\textwidth]{figures/moe_layerwise_routing_percentage_heatmap_step150000_ema.pdf}
  %   \caption{Contact map prediction comparison on CASP15 targets. We visualize predicted long-range contact maps from different pretrained encoders under the same supervised contact-head protocol. The comparison illustrates whether each model preserves
  %   structure-relevant nonlocal residue couplings after pretraining or adaptation.}
  %   \label{fig:contact_map_comparison}
  % \end{figure*}
 
To analyze expert specialization, we integrated three annotation sources: 19 curated Swiss-Prot residue- and region-level feature types, Pfam 38.2 scans covering 19,783 families and 844 clans, and official TAPE Q3/Q8 secondary-structure labels. Across categories spanning thousands to millions of residues, 80--99\% of residues in specific categories were routed to one expert at particular layers, as illustrated in Fig.~\ref{fig:model-analysis}a. To control for amino-acid and positional biases, we paired each annotated residue with an unannotated residue from the same protein, prioritizing identical amino acids and relative-position bins. Enrichment over these matched backgrounds is reported in Table~\ref{tab:moe-routing-enrichment}. To test whether these preferences were functionally relevant, we replaced Top-1 selections with alternative experts at selected layers and measured performance changes, as shown in Fig.~\ref{fig:model-analysis}b.

\begin{table}[t]
\centering
\caption{Expert routing across biological annotations. Pfam results are
restricted to clans with $n_{\mathrm{prot}}\geq500$}
\label{tab:moe-routing-enrichment}
\footnotesize
\setlength{\tabcolsep}{4pt}
\begin{tabular}{lrrr}
\toprule
Annotation & Classes & Residues & Target/Ref. (\%) \\
\midrule
Pfam clans
    & 196 & 71.29M & \textbf{92.5}/24.3 \\
Swiss-Prot sites
    & 19  & 39.14M & \textbf{64.2}/31.7 \\
TAPE Q3
    & 3   & 2.67M  & \textbf{60.7}/30.4 \\
\bottomrule
\end{tabular}
\end{table}

These results reveal a biological hierarchy in MoE routing across local structures, functional residues, and conserved domains. Rather than interchangeable branches, experts exhibit multiscale specialization supporting remote-homology detection, domain and functional-site annotation, secondary-structure prediction, and targeted adaptation.

% \begin{table}[t]
%   \centering
%   \caption{Efficiency comparison.}
%   \label{tab:efficiency}
%   \begin{tabular}{lcccc}
%     \hline
%     Model & Params & Throughput & Memory & Inference Cost \\
%     \hline
%     ESM2 continuation & TBD & TBD & TBD & TBD \\
%     ProtLingo-Memory & TBD & TBD & TBD & TBD \\
%     ProtLingo-MoE & TBD & TBD & TBD & TBD \\
%     ProtLingo & TBD & TBD & TBD & TBD \\
%     \hline
%   \end{tabular}
% \end{table}

\section{Discussion}
\paragraph{Limitations.} Although ProtLingo improves the compact ESM2 backbone across several settings, the gains over ESM2-150M are bounded by the lightweight nature of the current modules, suggesting that richer routing designs may further unlock conditional specialization;
  future work could explore alternative MoE routers such as hierarchical constrained routing~\cite{vashkelis2026himoehierarchicalinstanceconditionedmixtureofexperts}.
\paragraph{Conclusion.} We presented ProtLingo, an efficient protein language modeling framework that combines centered latent $N$-gram memory with sparse expert routing. By introducing reusable local-context memory and residue-dependent computation into a pretrained ESM2 backbone, ProtLingo improves mutation-sensitive modeling without relying on substantially larger dense models. Experiments across sequence-function prediction, contact prediction, and language modeling demonstrate strong performance at the 150M scale while preserving structure-relevant representations. These results highlight conditional memory and expert routing as effective alternatives to continued dense scaling.

%
% References. These lines must be placed at the end of your paper.
% Add your entries to aaai2027.bib and cite them with \cite / \citep / \citet.
\bibliography{aaai2027}

\newpage ~
\newpage
\section{Supplementary Details}
\label{sec:supplementary_details}

\subsection{Implementation Details}
\label{sec:implementation_details}

This section provides the implementation and training details needed for reproduction. Unless otherwise stated, layer indices are zero-based.

\subsubsection{Model Architecture}

We continued pretraining from the official pretrained ESM-2 150M checkpoint. The backbone contains 30 Transformer blocks with a hidden size of 640, 20 attention heads, and a feed-forward width of 2,560. The maximum input length was 1,024 tokens, including sequence boundary tokens. The tokenizer, vocabulary, positional encoding, and masked-language-model head followed the original ESM-2 configuration. Transformer blocks were implemented with NVIDIA Transformer Engine and trained in BF16.

\subsubsection{MoE Architecture}

Blocks 0--2 retained their original dense feed-forward networks, while blocks 3--29 were converted to MoE layers using the \texttt{shared\_routed\_split} implementation. In each converted block, the 2,560 intermediate channels of the pretrained feed-forward network were split into two complementary 1,280-channel partitions. One partition initialized an always-active shared expert, and the other initialized four routed experts.

For each token, the shared expert and one top-1 routed expert were activated, preserving an active feed-forward width of $1,280+1,280=2,560$. The router was initialized with a standard deviation of 0.01. The load-balancing and router z-loss coefficients were 0.01 and 0.001, respectively. No perturbation was added to the routed experts at initialization; expert specialization emerged during continued pretraining.

\subsubsection{LNgram Architecture}

LNgram modules were inserted before blocks 1, 11, and 21. All three modules used a centered three-token window, two bits per route, a memory dimension of 8, and a latent dimension of 256. With a hidden size of 640, each module contained 320 routes. Each route used four learned codes and one boundary code, yielding $5^3=125$ memory addresses for a three-token window.

Memory tables were zero-initialized, the residual scale was initialized to 1.0, and LNgram dropout was disabled. Counterfactual surrogate gradients were computed in chunks of 64 routes. For packed inputs, window construction and local convolution respected \texttt{cu\_seq\_lens}, preventing information flow across protein boundaries.

\subsubsection{Training Configuration}

Training used eight-way DistributedDataParallel on a single node without gradient accumulation. Protein sequences were packed  with a maximum length of 1,024. The per-GPU token budget was 98,304, corresponding to 786,432 tokens per optimizer step. The MLM corruption probability was 15\%.

Training used the NVIDIA ESM-2 UniRef pretraining data in Parquet format, with the corresponding held-out split used for validation. The model was continuously pretrained for 40,000 optimizer steps, and the step-40k checkpoint was used for the reported downstream evaluations.

\subsubsection{Optimization}

Two-dimensional weight matrices were optimized with Muon using a learning rate of $1.5\times10^{-3}$, momentum 0.95, five Newton--Schulz iterations, and Nesterov momentum. Embeddings, normalization parameters, biases, the MLM head, and router parameters were optimized with AdamW at $1\times10^{-4}$, using $\beta=(0.9,0.98)$ and $\epsilon=10^{-8}$. LNgram memory tables used an AdamW learning rate of $2\times10^{-4}$ without weight decay; the remaining parameter groups used a weight decay of 0.01.

We used a warmup--stable--decay schedule with 10,000 warmup steps. An FP32 exponential moving average with a decay of 0.9999 was updated after every optimizer step. The backbone and the custom MoE and LNgram paths used BF16; FP8 was not enabled for the reported experiment.

\subsubsection{Hardware and Software}

Experiments were based on the BioNeMo Recipes ESM-2 training path and NVIDIA Transformer Engine 2.10, and were run on a single eight-GPU H20 node with PyTorch.

%%%%%%%%%%%%%%%%%%%%%%%%%%%%%%%%%%%%%%%%%%%%%%%%%%%%%%%%%%%%%%%%%%%%%%%%%%%%%%%%%%%%%%%%%%%%%%%%%%%%%%%%%%%%%%%%%%%%%%%%%%%%%%%%%%%%%%%%%%%%%
% \subsection{Benchmark Protocols and Evaluation Metrics}
% \label{sec:metric_explanation}

% \subsubsection{Benchmark details}
% \subsubsection{MLM Loss and Perplexity}
% \subsubsection{ProteinGym DMS Spearman Correlation}
% \subsubsection{FLIP Metrics}
% \subsubsection{Contact Prediction Metrics}
% \subsubsection{Routing and Expert-Specialization Metrics}
\subsection{Benchmarking and Evaluation Metrics}
\label{sec:metric_explanation}

We evaluate protein language models on three benchmarks: the ProteinGym v1.3 DMS substitution benchmark~\cite{notin2023proteingym}, the GB1, AAV, and Meltome landscapes from FLIP~\cite{dallago2021flip}, and the CASP15 benchmark for supervised long-range contact prediction. Performance is measured using Spearman's rank correlation and area under the receiver operating characteristic curve (AUC) on ProteinGym, Spearman's rank correlation on FLIP, and P@L, P@L/2, and P@L/5 on CASP15.

\subsubsection{Benchmarking}

\paragraph{ProteinGym.} ProteinGym evaluates whether a protein language model can estimate the effects of amino-acid substitutions on experimentally measured functional phenotypes in the \textbf{zero-shot} setting. In a deep mutational scanning (DMS) assay, mutations are systematically introduced into the same protein, and the resulting variants are experimentally scored for a functional phenotype; each assay therefore provides a large sequence-to-phenotype test set. We use the ProteinGym v1.3 DMS substitution benchmark, which comprises 217 assays covering approximately 2.4 million experimentally measured variants. No DMS labels are used to train an assay-specific prediction head.

For each variant, we compute a masked-marginal score. At every mutated position, the corresponding residue in the wild-type sequence is masked, and the conditional log-probabilities assigned to the mutant and wild-type residues are compared:
\begin{equation}
s\!\left(x^{\mathrm{mut}}\right)=\sum_{i\in\mathcal{M}}\left[\log p_{\theta}\!\left(x_i^{\mathrm{mut}}\mid x_{\setminus i}^{\mathrm{wt}}\right)-\log p_{\theta}\!\left(x_i^{\mathrm{wt}}\mid x_{\setminus i}^{\mathrm{wt}}\right)\right],
\end{equation}
where $\mathcal{M}$ is the set of mutated positions and $x_{\setminus i}^{\mathrm{wt}}$ denotes the wild-type sequence with position $i$ masked. The score contains one log-probability difference for a single substitution and sums the sitewise differences for a multiple-substitution variant. When a sequence exceeds the model context length, a mutation-centered context window is selected separately for each queried position.

Spearman correlation and Area Under the Receiver Operating Characteristic Curve (AUC) are first computed independently within each assay. The reported ProteinGym scores use a nested aggregation procedure~\cite{notin2023proteingym}:
\begin{equation}
q_{\mathrm{ProteinGym}}=\frac{1}{|\mathcal{C}|}\sum_{c\in\mathcal{C}}\frac{1}{|U_c|}\sum_{u\in U_c}\frac{1}{|A_{u,c}|}\sum_{a\in A_{u,c}}q_a,
\end{equation}
where $q_a$ is the metric for assay $a$, $A_{u,c}$ is the set of assays associated with UniProt entry $u$ and functional category $c$, and $U_c$ is the set of UniProt entries in category $c$. The five categories are Activity, Binding, Expression, Organismal Fitness, and Stability. This procedure first averages repeated assays for the same protein and category, then averages proteins within each category, and finally assigns equal weight to the five category-level scores.

\paragraph{FLIP.} A protein fitness landscape maps protein sequences to experimentally measured functional values. FLIP evaluates how well pretrained protein representations transfer to \textbf{supervised fitness prediction} under train--test splits that probe generalization across mutations, fitness ranges, and protein families~\cite{dallago2021flip}. We evaluate five GB1 splits, seven AAV splits, and three Meltome splits. GB1 measures binding-related fitness, AAV measures capsid fitness, and Meltome measures protein thermostability.

We use a common frozen-representation probing procedure for every model. The encoder parameters remain fixed, and residue representations from the final hidden layer are mean-pooled after removing special tokens to obtain one vector per protein. Sequences exceeding the model context length are divided into non-overlapping windows. Residue embeddings are summed within each window and combined across windows using a residue-count-weighted mean, ensuring that every residue contributes equally to the resulting protein representation.

The pooled representations are standardized and used to fit a ridge-regression probe. The regularization coefficient is selected from $\{10^{-1},10^{0},10,10^{2},10^{3},10^{4}\}$ according to Spearman correlation on the marked validation examples when a split contains at least ten such examples; otherwise, it is selected by five-fold cross-validation on the training set. After hyperparameter selection, the validation examples are folded back into the training data, and the probe is refitted before evaluation on the held-out test set. Spearman correlation is computed separately for every split, and the reported GB1, AAV, and Meltome scores are the unweighted arithmetic means of the five, seven, and three split-level correlations, respectively.

\paragraph{CASP15.} A protein contact map identifies residue pairs that are spatially close in the folded three-dimensional structure. Long-range contact prediction excludes pairs that are nearby in the amino-acid sequence and therefore probes whether pretrained representations preserve nonlocal structural information. We evaluate the models on the CASP15 benchmark using the attention-based supervised contact-prediction setup adopted in MSA Transformer~\cite{rao2021msa}.

Each encoder is frozen, and a separate contact head is fitted for every model using the same head form, training data, and training procedure. Self-attention maps from all layers and attention heads are restricted to residue positions, symmetrized, and corrected using average product correction (APC). The resulting pairwise features are provided to an L1-regularized logistic-regression head, which produces an $L\times L$ matrix of contact probabilities for a protein of length $L$.

A residue pair is labeled as a contact when its C$\beta$--C$\beta$ distance is strictly below $8\,\mathrm{\AA}$, with C$\alpha$ used for glycine. Residue pairs with missing structural coordinates are excluded, and only long-range pairs satisfying $|i-j|\geq24$ are evaluated. One copy of each unordered residue pair is retained from the upper triangle of the contact map. Predictions are ranked by their contact probabilities, and P@L, P@L/2, and P@L/5 are computed for each CASP15 domain before being macro-averaged across domains.

\subsubsection{Evaluation Metrics}

\paragraph{Spearman's Rank Correlation.} Spearman's rank correlation coefficient measures the agreement between the rankings induced by model predictions and experimentally measured values:
\begin{equation}
\rho=\operatorname{corr}\!\left(\operatorname{rank}(\mathbf{y}),\operatorname{rank}(\hat{\mathbf{y}})\right),
\end{equation}
where $\mathbf{y}$ and $\hat{\mathbf{y}}$ denote the experimental measurements and model predictions, respectively. ProteinGym adopts Spearman correlation as its primary metric for continuous DMS measurements because the relationship between an experimentally measured protein phenotype and biological fitness may be nonlinear~\cite{notin2023proteingym}. We compute $\rho$ independently for each ProteinGym assay before applying the nested aggregation procedure described above. For FLIP, Spearman correlation is computed on each held-out test split, and the reported GB1, AAV, and Meltome values are the unweighted arithmetic means of their corresponding split-level correlations~\cite{dallago2021flip}. Higher values indicate better agreement between the predicted and experimental rankings.

\paragraph{Area Under the Receiver Operating Characteristic Curve(AUC).} AUC measures how well continuous model scores distinguish experimentally favorable variants from unfavorable variants using the assay-specific binary labels supplied by ProteinGym~\cite{notin2023proteingym}. It can be interpreted as the probability that a randomly selected positive variant receives a higher model score than a randomly selected negative variant:
\begin{equation}
\mathrm{AUC}=\Pr\!\left(\hat{y}^{+}>\hat{y}^{-}\right)+\frac{1}{2}\Pr\!\left(\hat{y}^{+}=\hat{y}^{-}\right),
\end{equation}
where $\hat{y}^{+}$ and $\hat{y}^{-}$ denote model scores for positive and negative variants, respectively. ProteinGym includes AUC as a complementary metric because rank correlation may be less informative for assays with bimodal experimental measurements. AUC is computed independently for every assay containing both binary classes and is then combined using the same nested ProteinGym aggregation procedure as Spearman correlation. Higher AUC indicates better discrimination, while a value of $0.5$ corresponds to random ranking.

\paragraph{Long-Range Precision at $K$.} Long-range contact prediction is evaluated using precision among the highest-scoring residue pairs, following the contact-prediction setting used in MSA Transformer~\cite{rao2021msa}. For a protein domain of length $L$, let $\mathcal{V}$ denote the set of structurally resolved residue pairs satisfying $|i-j|\geq24$. After ranking these pairs by their predicted contact probabilities, precision at $K$ is defined as
\begin{equation}
\mathrm{P@K}=\frac{1}{K}\sum_{(i,j)\in\operatorname{TopK}(\mathcal{V})}\mathbf{1}\!\left[d_{ij}<8\,\mathrm{\AA}\right],
\end{equation}
where $d_{ij}$ is the C$\beta$--C$\beta$ distance between residues $i$ and $j$, with C$\alpha$ used for glycine. We evaluate $K=L$, $\lfloor L/2\rfloor$, and $\lfloor L/5\rfloor$, corresponding to P@L, P@L/2, and P@L/5, respectively; when fewer than $K$ valid residue pairs are available, $K$ is capped by the number of valid pairs. Each metric is computed independently for every CASP15 domain and then macro-averaged across domains.
%%%%%%%%%%%%%%%%%%%%%%%%%%%%%%%%%%%%%%%%%%%%%%%%%%%%%%%%%%%%%%%%%%%%%%%%%%%%%%%%%%%%%%%%%%%%
\subsection{Ablation Studies}
\label{sec:ablation_studies}

In the main paper, we discussed the effectiveness of the principal components of our model, including sparse expert routing, local motif memory, and data sampling. Here, we provide further motivation for these design choices and present the corresponding ablation analyses.

\subsubsection{Why Use This MoE Design and Expert Configuration?}

Protein sequences contain segments with markedly different physicochemical and functional properties. For example, transmembrane regions are typically enriched in hydrophobic residues, whereas solvent-exposed loops and disordered regions contain more hydrophilic or charged residues. Catalytic sites and conserved motifs impose yet another type of local constraint through specific residue combinations. Processing all these contexts with the same feed-forward parameters may require a single dense module to learn several competing transformations.

MoE provides a form of conditional computation that is well suited to such heterogeneous sequence contexts. Given its contextualized representation, each token can activate a different routed expert, allowing different parameter subsets to specialize according to recurring contextual patterns. This specialization is not assumed to be a strict one-to-one mapping between experts and biological classes; rather, it emerges from training as differences in routing preference and feature enrichment.

We retain the first three Transformer blocks as dense layers, as early layers are generally understood to encode lower-level residue information before higher-level contextual representations emerge. Blocks 3--29 are converted to MoE layers, where more contextualized representations can support meaningful expert specialization. Each MoE layer contains one always-active shared expert and several routed experts, with one or more routed expert selected for each token. The shared expert is intended to preserve common protein-language transformations, while the routed experts provide context-dependent capacity without increasing the active feed-forward width.

\subsubsection{Why Use the Proposed LNgram Design?}

Short amino-acid motifs encode important biological signals. For example, N-X-S/T represents a potential N-linked glycosylation site, while related local patterns occur in cleavage sites, binding regions, transmembrane segments, and conserved catalytic motifs. Although self-attention can model such patterns implicitly, it must repeatedly reconstruct them through token interactions.

LNgram instead implements an explicit, learnable lookup table. Learned routers convert local hidden states into discrete codes, whose combination forms a key for retrieving a trainable memory vector. The router and memory entries are optimized end-to-end with the MLM objective, so the table learns functional motif representations rather than fixed occurrence statistics.

More generally, LNgram uses a centered \(2n+1\)-token window containing the target residue and \(n\) residues on each side. This odd-length design provides a unique center and symmetric context. Three-token windows capture immediate dependencies such as N-X-S/T, whereas five- or seven-token windows can represent longer functional or structural patterns. However, increasing \(n\) expands the lookup space exponentially, reducing address coverage and increasing memory cost and sparsity. We therefore use centered three-token windows as a balance between biological context, statistical coverage, and efficiency.

LNgram modules are inserted at layers 1, 11, and 21 to retrieve motif information from residue-level, intermediate, and contextualized representations. Each module concatenates 320 route-specific 8-dimensional memory vectors into a 2,560-dimensional representation, which is compressed to a 256-dimensional latent space before projection back to the model. This bottleneck reduces parameters and computation, filters redundant or noisy route features, and prevents the memory branch from overwhelming the pretrained representation.

Removing LNgram degrades both MLM perplexity and downstream mutation-effect prediction, indicating that the learnable motif memory provides information complementary to the Transformer backbone and MoE routing.

\subsubsection{Why Use an 80/20 Mixed Sampling Strategy?}

Three straightforward sampling strategies have different potential limitations. First, UniRef90 row-uniform sampling may overrepresent large and highly redundant protein families. Second, UniRef50 representative sampling, which uses one fixed sequence from each UniRef50 cluster, improves family-level balance but removes most within-family variation. Third, UniRef50-to-UniRef90 hierarchical sampling first selects a UniRef50 cluster uniformly and then randomly selects one of its UniRef90 members. This restores some within-family diversity but may distort the natural family-size distribution and repeatedly oversample members of small clusters.

We therefore use a mixture of 80\% UniRef50 representative sampling and 20\% UniRef90 row-uniform sampling. The UniRef50 component is intended to provide broad family-level coverage, while the UniRef90 component partially restores natural sequence abundance and within-family variation. This design should be regarded as a practical compromise and a working hypothesis, since the individual strategies and the 80/20 ratio have not yet been validated through a fully controlled ablation.

%%%%%%%%%%%%%%%%%%%%%%%%%%%%%%%%%%%%%%%%%%%%%%%%%%%%%%%%%%%%%%%%%%%%%%%%%%%%%%%%%%%%%%%%%%%%%%%%%%%%%

\subsection{Baselines}

We compare ProtLingo with six protein language models: ESM-1b, ESM2-150M, ESM2-650M, ProtBert, CARP-640M, and RITA XL. The three ESM models are the closest points of comparison because ProtLingo is initialized from ESM2-150M, while the remaining models extend the comparison to protein language models developed with different architectures and pretraining objectives.

\paragraph{ESM family.}
The ESM baselines cover two generations of bidirectional Transformer protein language models trained with masked language modeling \cite{rives2021biological,linEvolutionaryscalePredictionAtomiclevel2023}. ESM-1b is a 650M-parameter model from the original ESM series, whereas ESM2-150M and ESM2-650M belong to the subsequent ESM2 generation. Among them, ESM2-150M is the most direct baseline for ProtLingo: it provides the pretrained checkpoint from which our model is initialized, and its 150M dense parameters closely match the 153M parameters activated by ProtLingo. ESM2-650M places this matched-backbone comparison alongside conventional dense scaling, with ESM-1b adding a similarly sized model from the preceding generation.

\paragraph{ProtBert.}
We also include ProtBert, a 420M-parameter BERT-style protein language model introduced in the ProtTrans framework \cite{elnaggarProtTransCrackingLanguage2021}. ProtBert was pretrained on UniRef100 using masked language modeling. Although it follows the same broad bidirectional pretraining formulation as ESM, it was developed through a separate model and pretraining pipeline, adding an intermediate-scale dense Transformer outside the ESM family.

\paragraph{CARP-640M.}
CARP-640M retains masked language modeling but replaces Transformer self-attention with a ByteNet-based architecture built from dilated convolutions \cite{yang2024convolutions}. The model was pretrained on UniRef50 and contains approximately 640M parameters, placing it at a similar scale to ESM-1b and ESM2-650M while introducing a substantially different sequence-modeling architecture.

\paragraph{RITA XL.}
RITA XL differs from the preceding baselines in both scale and pretraining formulation \cite{hesslowRITAStudyScaling2022}. It is a 1.2B-parameter decoder-only Transformer trained autoregressively on more than 280 million protein sequences from UniRef100, with each amino acid predicted from the preceding sequence context. RITA XL is therefore the only causal language model and the largest dense baseline included in our comparison.
%%%%%%%%%%%%%%%%%%%%%%%%%%%%%%%%%%%%%%%%%%%%%%%%%%%%%%%%%%%%%%%%%%%%%%%%%%%%%%%%%%%%%%%%%%%%%%%%%%%%

\subsection{Additional Results}
\label{sec:additional_results}

\subsubsection{Layer-resolved Expert-routing Profiles}
\label{sec:layer_resolved_routing}

The main text summarizes peak expert enrichment across biological annotations and presents representative examples (Fig.~\ref{fig:model-analysis}a and Table~\ref{tab:moe-routing-enrichment}). Here, we provide the corresponding layer-resolved routing profiles to show how these preferences vary across the 27 MoE blocks.

\begin{figure*}[t]
    \centering
    \includegraphics[width=1.0\textwidth]
    {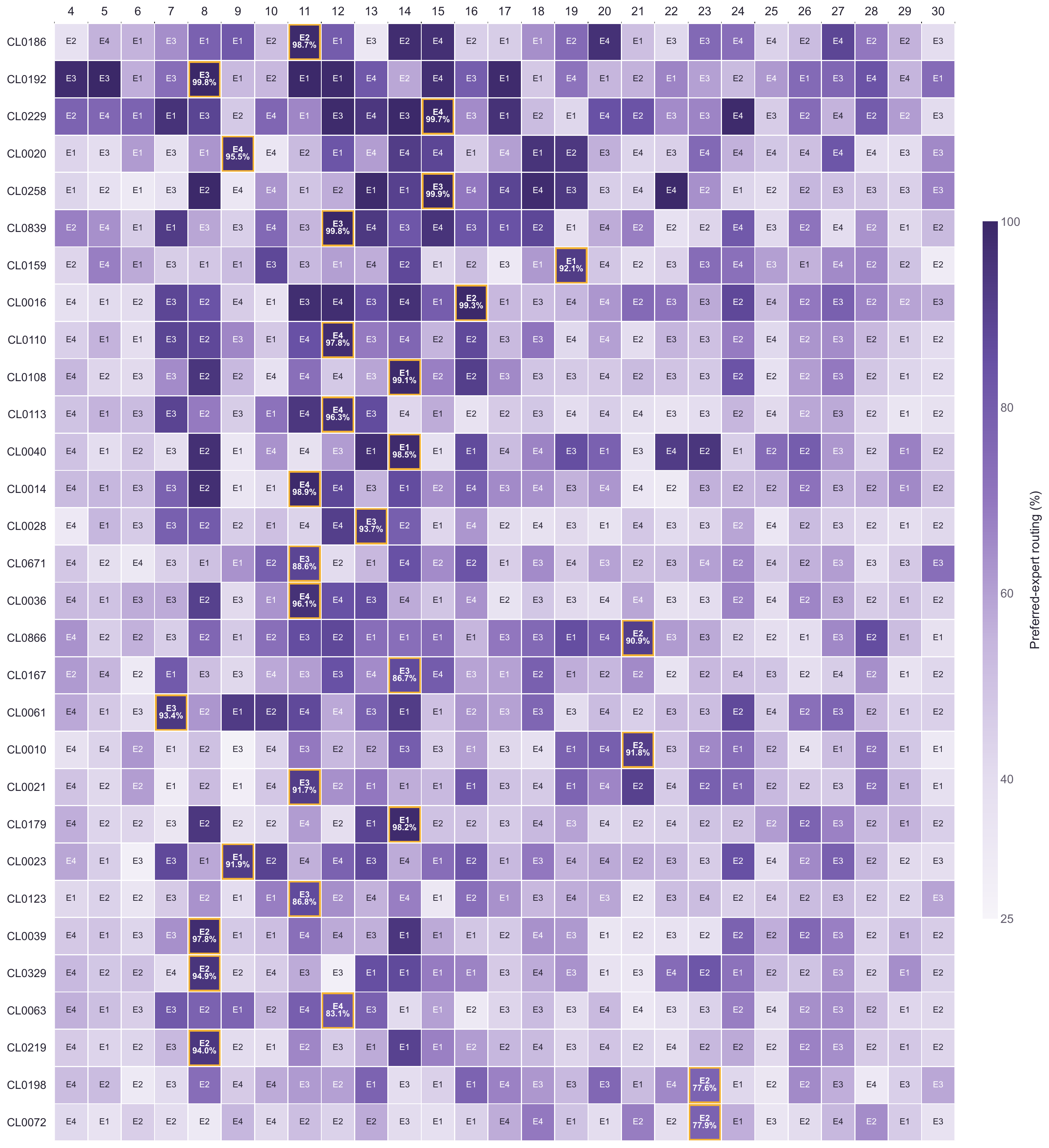}
    \caption{
    \textbf{Layer-resolved routing profiles of high-support Pfam clans.}
    The analysis includes 30 clans represented by at least 2,500 proteins. Each cell identifies the expert receiving the largest fraction of clan residues at the corresponding MoE block; color intensity denotes that fraction. Gold outlines indicate the maximum routing concentration across blocks for each clan.}
    \label{fig:moe_all_layers_clans30}
\end{figure*}

Figure~\ref{fig:moe_all_layers_clans30} reports the complete profiles for 30 high-support Pfam clans represented by at least 2,500 proteins. Routing concentration is distributed across different depths, and the identity of the preferred expert frequently changes between blocks. Thus, the peak-layer enrichments reported in the main text are not attributable to a single globally dominant expert or a single MoE block.

\begin{figure*}[t]
    \centering
    \includegraphics[width=1.0\textwidth]
    {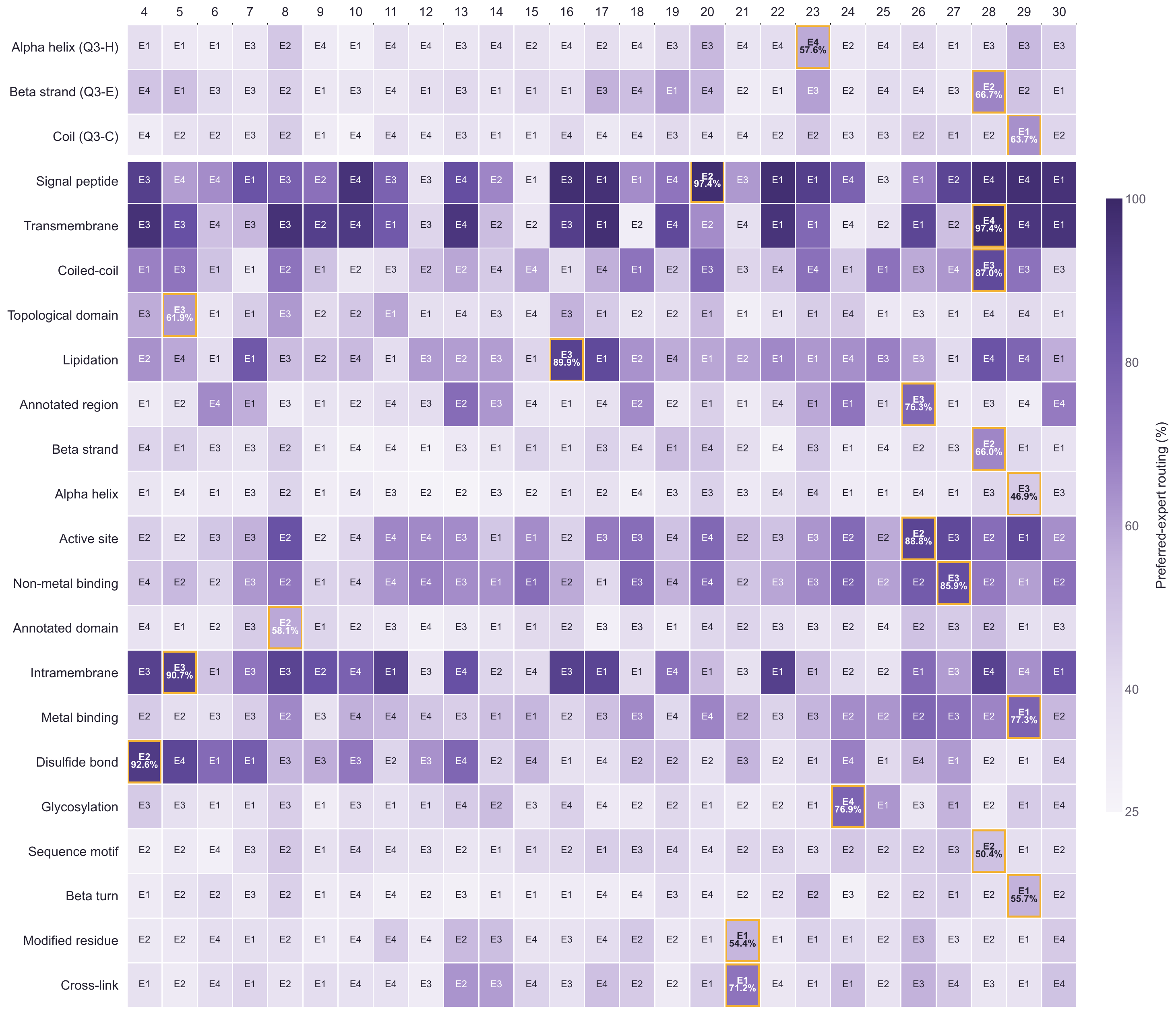}
    \caption{
    \textbf{Layer-resolved routing profiles of structural and functional
    annotations.}
    Rows comprise the three TAPE Q3 secondary-structure classes and 19 Swiss-Prot residue- or region-level annotations. Each cell dentifies the preferred expert at the corresponding MoE block, with color intensity indicating its routing fraction. Gold outlines mark the maximum routing concentration across blocks for each annotation.}
    \label{fig:moe_all_layers_q3_sites22}
\end{figure*}

Figure~\ref{fig:moe_all_layers_q3_sites22} provides the same analysis for the three Q3 secondary-structure classes and 19 Swiss-Prot annotations. The Q3 classes generally exhibit more distributed routing, whereas several functional and topological annotations display sharper, depth-specific preferences. These profiles complement the aggregate enrichment statistics by preserving the layer at which each routing preference emerges.

Many Pfam clans and several covalent or topological annotations peak in early or intermediate blocks, whereas the Q3 classes and several catalytic, binding, and metal-associated annotations peak at greater depths. This pattern may reflect a progression from local sequence and family-scaffold recognition to representations integrating broader structural and functional context.

\end{document}